\PassOptionsToPackage{dvipsnames}{xcolor}
\PassOptionsToPackage{table}{xcolor}
\documentclass[11pt,a4paper]{article}
\usepackage{times,latexsym}
\usepackage{url}
\usepackage[T1]{fontenc}
\usepackage{booktabs}
\usepackage{amsmath}
\usepackage{ amssymb }
\usepackage{graphicx}
\usepackage{subcaption}
\usepackage{linguex}
\usepackage{qtree}
\usepackage{tabularx}
\usepackage{multicol}
 \usepackage{vwcol}

\usepackage[table]{xcolor}
\usepackage{collcell}
\usepackage{hhline}
\usepackage{pgf}
\usepackage{multirow}
\usepackage{pifont}

\usepackage{scalerel}
\def\noattn{\scalerel*{\includegraphics{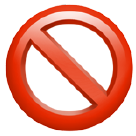}}{X}}
\def\locattn{\scalerel*{\includegraphics{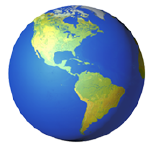}}{X}}
\def\contattn{\scalerel*{\includegraphics{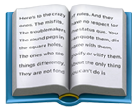}}{X}}

\def\colorModel{rgb} 
\newcommand\ColCell[1]{
  \pgfmathparse{#1<1.2?1:0}  
    \ifnum\pgfmathresult=0\relax\color{white}\fi
  \pgfmathsetmacro\compA{1 - 0.26 * #1)} 
  \pgfmathsetmacro\compB{1 - 0.32 * #1)} 
  \pgfmathsetmacro\compC{1 - 0.17 * #1)} 
  \edef\x{\noexpand\centering\noexpand\cellcolor[\colorModel]{\compA,\compB,\compC}}\x #1
  } 
\newcolumntype{E}{>{\collectcell\ColCell}m{1.3cm}<{\endcollectcell}}  
\newcolumntype{F}{>{\collectcell\ColCell}m{1.7cm}<{\endcollectcell}}  
\newcolumntype{Y}{>{\collectcell\ColCell}m{2.1cm}<{\endcollectcell}}  

\newcommand\ColCellB[1]{
  \pgfmathparse{#1<1.2?1:0}  
    \ifnum\pgfmathresult=0\relax\color{white}\fi
    \pgfmathsetmacro\compA{1 - 0.50 * #1)} 
  \pgfmathsetmacro\compB{1 - 0.21 * #1)} 
  \pgfmathsetmacro\compC{1 - 0.50 * #1)} 
  \edef\x{\noexpand\centering\noexpand\cellcolor[\colorModel]{\compA,\compB,\compC}}\x #1
  } 
\newcolumntype{G}{>{\collectcell\ColCellB}m{1.3cm}<{\endcollectcell}}  
\newcolumntype{Q}{>{\collectcell\ColCellB}m{0.5cm}<{\endcollectcell}}  
\newcolumntype{H}{>{\collectcell\ColCellB}m{1.7cm}<{\endcollectcell}}  
\newcolumntype{Z}{>{\collectcell\ColCellB}m{2.1cm}<{\endcollectcell}}  

\newcolumntype{O}[1]{>{\centering\arraybackslash\hspace{0pt}}p{#1}}

\usepackage[acceptedWithA]{tacl2018v2}
 \usepackage{cleveref}[2012/02/15]

\usepackage{xspace,mfirstuc,tabulary}

\newif\iftaclinstructions
\taclinstructionsfalse 
\iftaclinstructions
\renewcommand{\confidential}{}
\renewcommand{\anonsubtext}{(No author info supplied here, for consistency with
TACL-submission anonymization requirements)}
\newcommand{\instr}
\fi

\iftaclpubformat 

\else

\fi

\title{Does syntax need to grow on trees?\\  Sources of hierarchical inductive bias in sequence-to-sequence networks}

\author{
 R.\ Thomas McCoy \\
 Department of Cognitive Science \\
 Johns Hopkins University \\
  {\tt tom.mccoy@jhu.edu} \\
  \And
  Robert Frank \\
 Department of Linguistics \\
 Yale University \\
  {\tt robert.frank@yale.edu} \\
  \And
  Tal Linzen \\
 Department of Cognitive Science \\
 Johns Hopkins University \\
  {\tt tal.linzen@jhu.edu} \\
}

\date{}

\crefformat{footnote}{#2\footnotemark[#1]#3}

\hypersetup{draft} 
\begin{document}
\maketitle
\begin{abstract}
 Learners that are exposed to the same training data might generalize differently due to differing inductive biases.
 In neural network models, inductive biases could in theory arise from any aspect of the model architecture.
 We investigate which architectural factors 
 affect the generalization behavior of neural sequence-to-sequence models trained on two syntactic tasks, English question formation and English tense reinflection. 
 For both tasks, the training set is consistent with a generalization based on hierarchical structure and a generalization based on linear order.
 All architectural factors that we investigated qualitatively affected how models generalized, including factors with no clear connection to hierarchical structure. For example, LSTMs and GRUs displayed qualitatively different inductive biases. However, the only factor that consistently contributed a hierarchical bias across tasks was the use of a tree-structured model rather than a model with sequential recurrence, suggesting that human-like syntactic generalization requires architectural syntactic structure. 
\end{abstract}

\setlength{\Exlabelwidth}{0.5em}
\setlength{\SubExleftmargin}{1.35em}

\setlength{\abovedisplayskip}{0pt}

\iftaclpubformat
\fi
\section{Introduction}

Any finite training set is consistent with multiple 
generalizations. Therefore, the way that a learner generalizes to unseen examples depends not only on the training data but also on properties of the learner.
Suppose a learner is told that a blue triangle is an example of a \textit{blick}. A learner preferring shape-based generalizations would conclude that \textit{blick} means ``triangle,'' while a learner preferring color-based generalizations would conclude that \textit{blick} means ``blue object'' \cite{landau1988importance}.
Factors that guide a learner to choose one generalization over another are called \textbf{inductive biases}.

\begin{figure}
    \centering
\begin{minipage}[t]{0.21\textwidth}
\raggedright\textbf{\textsc{move-main}:} Move the \underline{main verb's} auxiliary to the front of the sentence. \label{ex:hypa}
\raggedright\includegraphics[trim={0.3cm 1.5cm 0.5cm 1cm}, clip, width=\textwidth]{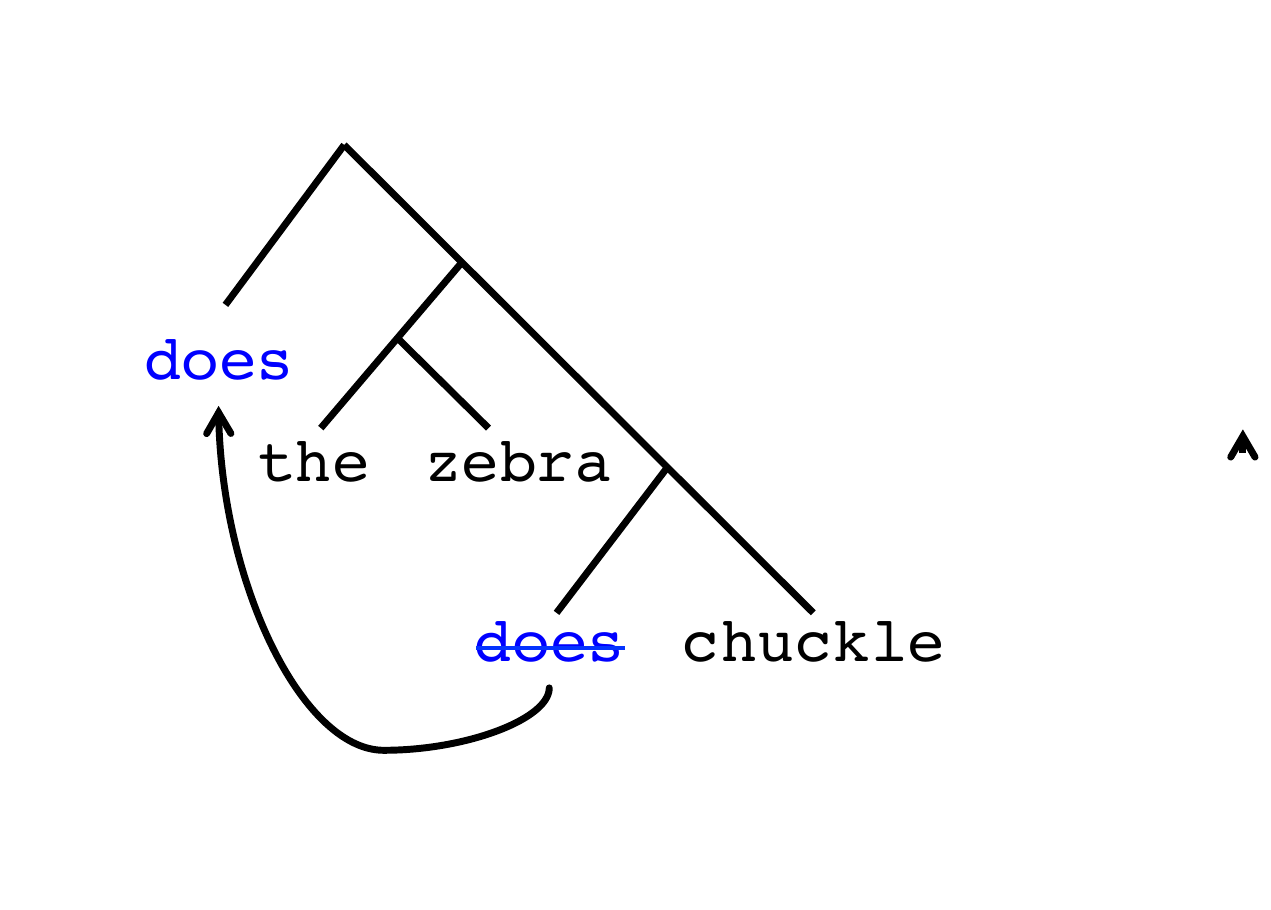}
\end{minipage}\hspace{0.02\textwidth}%
\begin{minipage}[t]{0.21\textwidth}
\textbf{\textsc{move-first}:} Move the \underline{\smash{linearly first}} auxiliary to the front of the sentence.\label{ex:hypb}
\raggedright\includegraphics[trim={0.3cm 1.5cm 0.5cm 2cm}, clip, width=\textwidth]{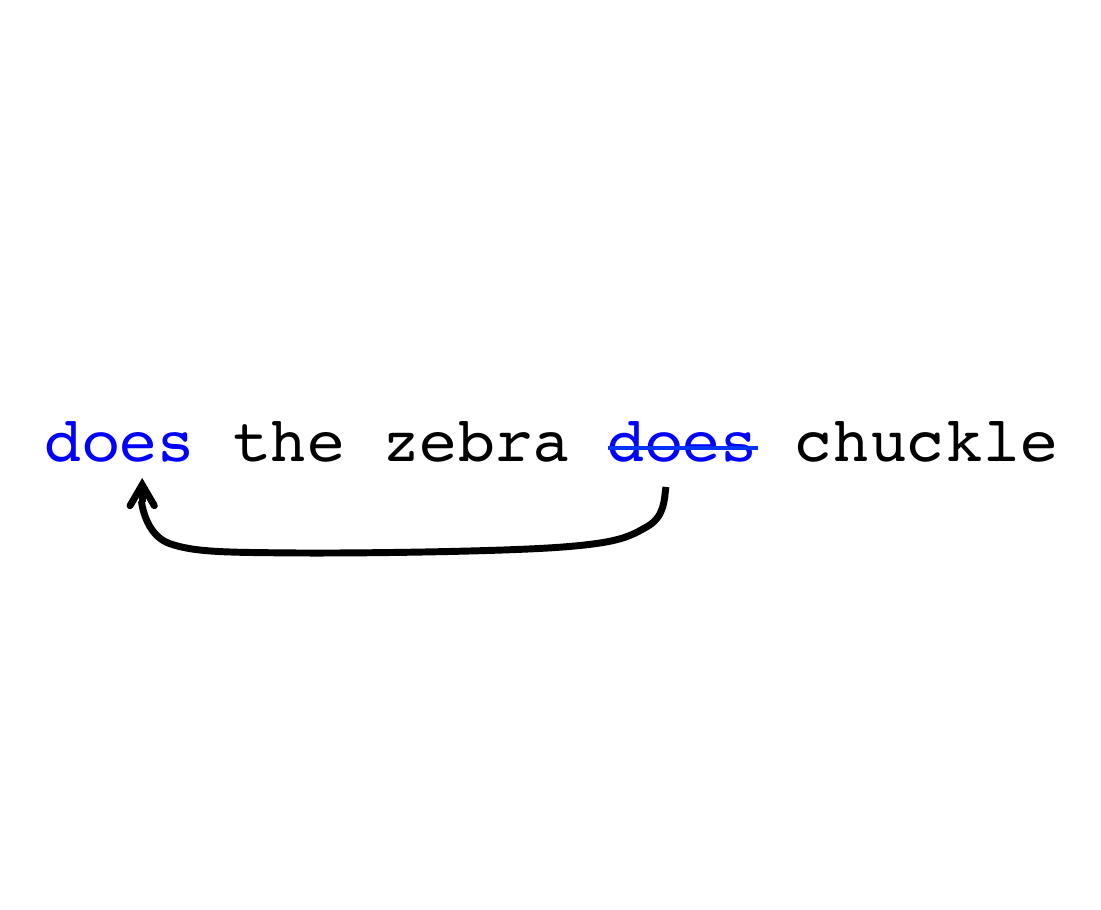}
\end{minipage}
    \caption{Two potential rules for English question formation.}
    \label{fig:rules}
\end{figure}

What properties of a learner cause it to have a particular inductive bias? 
We investigate this question with respect to sequence-to-sequence neural networks \cite{botvinick2006short, sutskever2014sequence}. 
As a test case for studying differences in how models generalize,  we use the syntactic task of \textbf{English question formation}, such as transforming \ref{ex:introa} into \ref{ex:introb}:

\ex. \label{ex:intro} \a. The zebra \textcolor{ForestGreen}{\textbf{does}} chuckle. \label{ex:introa}
\b. \textcolor{ForestGreen}{\textbf{Does}} the zebra chuckle? \label{ex:introb}

\noindent
Following \citeauthor{chomsky1980}'s (\citeyear{chomsky1980}) empirical claims about children's linguistic input, we constrain our training set to be consistent with two possible rules illustrated in Figure \ref{fig:rules}: \textsc{move-main} (a rule based on hierarchical syntactic structure) and \textsc{move-first} (a rule based on linear order). We then evaluate each trained model on examples where the rules make different predictions, such as \ref{ex:q2}: given \ref{ex:q2a}, \textsc{move-main} would generate \ref{ex:q2b} while \textsc{move-first} would generate \ref{ex:q2c}:

\ex. \label{ex:q2} \a. Your zebras that \textcolor{blue}{\textbf{don't}} dance \textcolor{ForestGreen}{\textbf{do}} chuckle. \label{ex:q2a}
\b. \textcolor{ForestGreen}{\textbf{Do}} your zebras that \textcolor{blue}{\textbf{don't}} dance chuckle?\label{ex:q2b}
\c. \textcolor{blue}{\textbf{Don't}} your zebras that dance \textcolor{ForestGreen}{\textbf{do}} chuckle?\label{ex:q2c}

Since no such examples appear in the training set, a model's behavior on them reveals which rule the model is biased toward. 
This task allows us to study a particular bias, namely a bias for hierarchical generalization, which is important for models of language because it has been argued to underlie human language acquisition \cite{chomsky1965}.

To test which models have a hierarchical bias, we use the question formation task and a second task: \textbf{tense reinflection}. For both tasks, our training set is ambiguous between a hierarchical generalization and a linear generalization. 
If a model chooses the hierarchical generalization for only one task, this preference is likely due to task-specific factors rather than a general hierarchical bias. On the other hand, 
a consistent preference for hierarchical generalizations across tasks 
would provide converging evidence that a model has a hierarchical bias. 

We find that all the factors we tested can qualitatively affect how a model generalizes on the question formation task. These factors are the type of recurrent unit, the type of attention, and the choice of sequential vs.\ tree-based model structure.
Even though all these factors affected the model's decision between \textsc{move-main} and \textsc{move-first},
only the use of a tree-based model can be said to impart a hierarchical bias, 
since this was the only model type that chose a hierarchical generalization across both of our tasks.
Specific findings that support these general conclusions include:

\begin{itemize}
    \item Generalization behavior is profoundly affected by the type of recurrent unit and the type of attention, and also by the \textit{interactions} between these factors.
    \item LSTMs and GRUs have qualitatively different inductive biases. The difference appears at least partly due to the fact that the values in GRU hidden states are bounded within a particular interval \cite{weiss2018}.
    \item Only a model built around the correct tree structure displayed a robust hierarchical bias across tasks. Sequentially-structured models failed to generalize hierarchically even when the input contained explicit marking of each sentence's hierarchical structure.
\end{itemize}

\noindent
Overall, we conclude that many factors can qualitatively affect a model's inductive biases, but human-like syntactic generalization may require specific types of high-level structure, at least when learning from text alone.

\begin{figure*}[!h]
\begin{center} 
\begin{tabular}{lp{7cm}lp{5cm}}
\fbox{\textcolor{white}{T}} & Training set, validation set, test set &
\fcolorbox{black}{gray!25}{\textcolor{gray!25}{T}}& Generalization set 
\end{tabular}

\vskip 0.03in

\begin{tabularx} {\textwidth} {p{1.1cm} | X | X} \hline
& \centering \textsc{decl} & \centering \textsc{quest}\arraybackslash \\ \hline
No RC & the newts do see my yak by the zebra . \newline $\rightarrow$ the newts do see my yak by the zebra . & the newts do see my yak by the zebra . \newline $\rightarrow$ do the newts see my yak by the zebra ? \\  \hline
\raggedright RC on object &  the newts do see my yak who does fly . \newline $\rightarrow$ the newts do see my yak who does fly . &   the newts do see my yak who does fly . \newline $\rightarrow$ do the newts see my yak who does fly ? \\ \hline
\raggedright RC on subject & the newts who don't fly do see my yak . \newline $\rightarrow$ the newts who don't fly do see my yak . & \cellcolor{gray!25}{ the newts who don't fly do see my yak . \newline $\rightarrow$ do the newts who don't fly see my yak ?}\\ \hline
\end{tabularx}
\end{center}
    \caption{The difference between the training set and generalization set. To save space, this table uses some words not present in the vocabulary used to generate the examples. RC stands for ``relative clause.''}
\label{table:typetask}

\end{figure*}

\section{The question formation task}\label{sec:qftask}

\subsection{Background}

The classic discussion of the acquisition of English question formation begins with two empirical claims: (i) disambiguating examples such as \ref{ex:q2} rarely 
occur in a child's linguistic input, but (ii) all learners of English nevertheless acquire \textsc{move-main} rather than \textsc{move-first}.
\newcite{chomsky1965,chomsky1980} uses these points to argue that humans must have an innate bias toward learning syntactic rules that are based on hierarchy rather than linear order (this argument is known as the \textit{argument from the poverty of the stimulus}).

There has been a long debate about this line of argument. Though some have discussed the validity of Chomsky's empirical claims \cite{crain1987structure,ambridge2008structure,pullum2002empirical,legate2002empirical}, most of the debate has been about which mechanisms could explain the preference for \textsc{move-main}.
These mechanisms include an assumption of substitutability \cite{clark2007polynomial}, a bias for simplicity \cite{perfors2011learnability}, exploitation of statistical patterns \cite{lewis2001learnability,reali2005uncovering}, and semantic knowledge \cite{fitz2017meaningful}; see \newcite{clark2010linguistic} for in-depth discussion.

These past works focus on the \textit{content} of the bias that favors \textsc{move-main} (i.e., which types of generalizations the bias supports), but we instead focus on the \textit{source} of this bias (i.e., which factors of the learner give rise to the bias). In the book \textit{Rethinking Innateness}, \newcite{elman1998rethinking} argue that innate biases in humans must arise from architectural constraints on the neural connections in the brain rather than from constraints stated at the symbolic level, under the assumption that symbolic constraints are unlikely to be specified in the genome. 
Here we use artificial neural networks to investigate whether syntactic inductive biases can emerge from architectural constraints.

\subsection{Framing of the task}

Following \newcite{frank2007} and \newcite{mccoy2018revisiting}, we train models to take a declarative sentence as input and to either output the same sentence unchanged, or transform that sentence into a question. The sentences were generated from a context-free grammar containing only the sentence types shown in Figure \ref{table:typetask} and using a 75-word vocabulary; the full grammar is at the project website.\cref{fn:website}
The different types of sentences vary in the linear position of the main auxiliary, such that a model cannot identify the main auxiliary with a simple positional heuristic. The task to be performed is indicated by the final input token, as in \ref{ex:decl} and \ref{ex:quest}:

\ex. \label{ex:decl} \a. \textit{Input:}\hfill your zebra does read . \textsc{decl}\phantom{XX}
\b. \textit{Output:} \hfill your zebra does read .  \phantom{\textsc{decl}XX}

\ex. \label{ex:quest} \a. \textit{Input:}\hfill your zebra does read . \textsc{quest}\phantom{X}
\b. \textit{Output:}\hfill does your zebra read ? \phantom{\textsc{quest}X}

During training, all question formation examples are consistent with both \textsc{move-first} and \textsc{move-main}, such that there is no direct evidence favoring one rule over the other (see Figure \ref{table:typetask}). 

To assess how models generalize, we evaluate them on a generalization set consisting of 
examples where \textsc{move-main} and \textsc{move-first} make different predictions due to the presence of a relative clause on the subject (see sentence \ref{ex:q2a}).

\subsection{Evaluation metrics}

We focus on two metrics. The first is \textbf{full-sentence accuracy on the test set}. That is, for examples drawn from the same distribution as the training set, does the model get the output exactly right?

For testing generalization to the withheld example type, a natural metric would be full-sentence accuracy on the generalization set. However, in preliminary experiments we found that most models rarely produced the exact output predicted by either \textsc{move-main} or \textsc{move-first}, as they tend to truncate the output, confuse similar words, and make other extraneous errors.
To abstract away from such errors, we use \textbf{first-word accuracy on the generalization set}. With both \textsc{move-first} and \textsc{move-main}, the first word of the question is the auxiliary that has been moved from within the sentence. If the auxiliaries in the relative and main clauses are distinct, this word alone is sufficient to differentiate the two rules. For example, in the bottom right cell of Figure \ref{table:typetask}, \textsc{move-main} predicts having \textit{do} at the start, while \textsc{move-first} predicts \textit{don't}.\footnote{We exclude from the generalization set cases where the two auxiliaries are the same.
We also exclude cases where one auxiliary is singular and the other plural so that a model cannot succeed by using heuristics based on the grammatical number of the subject.}
Models almost always produced either the main auxiliary or the first auxiliary as the first word of the output (over 98\% of the time for most models\footnote{The one exception is noted in the caption to Figure \ref{tab:architectures_final}.}), so a low first-word accuracy can be interpreted as high consistency with \textsc{move-first}.

\subsection{Architecture}

\begin{figure*}
    \centering
    \includegraphics[width=\textwidth]{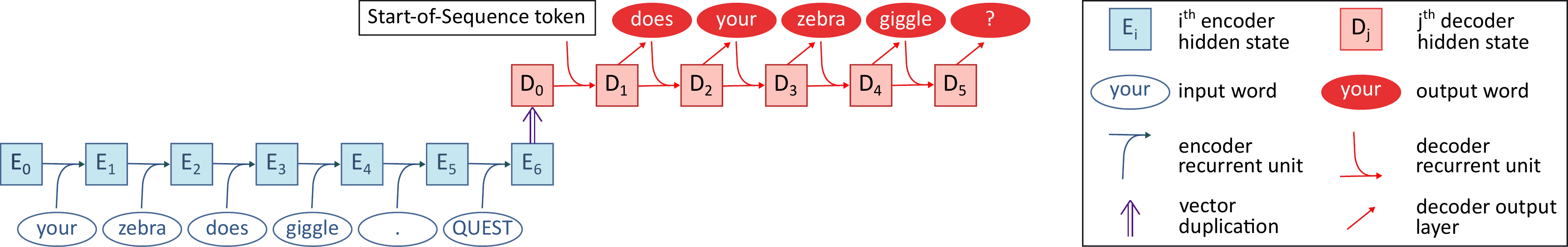}
    \caption{Sequential sequence-to-sequence model.}
    \label{fig:seq2seq}
\end{figure*}

We used the sequence-to-sequence architecture in Figure \ref{fig:seq2seq} \cite{sutskever2014sequence}.
This model consists of two neural networks: the \textbf{encoder} and the \textbf{decoder}.
The encoder is fed the input sentence one word at a time; after each word, the encoder updates its \textbf{hidden state}, a vector representation of the information encountered so far. 
After the encoder has been fed the entire input, its final hidden state ($E_6$ in Figure~\ref{fig:seq2seq}) is fed to the decoder, which generates an output sequence one word at a time based on its own hidden state, which is updated after each output word.
The weights that the encoder and decoder use to update their hidden states and generate outputs are learned via gradient descent; for more details, see Appendix \ref{app:details}.

\newcommand\noncolor{gray}
\begin{figure*}[]
    \centering
    \begin{subfigure}{0.5\textwidth}
    \centering
    \begin{tabular}{cEEE}
         &  \multicolumn{1}{c}{\noattn{}} & \multicolumn{1}{c}{\locattn{}} & \multicolumn{1}{c}{\contattn{}} \\
        SRN & 0.00 & 0.93 & 1.00 \\
        GRU & 0.88 & 0.77 & 1.00 \\
        LSTM & 0.94 & 0.98 & 1.00 \\
    \end{tabular}
    \subcaption{Full-sentence accuracy on the test set}
    \end{subfigure}%
    \begin{subfigure}{0.5\textwidth}
    \centering
    \begin{tabular}{cGGG}
         &  \multicolumn{1}{c}{\noattn{}} & \multicolumn{1}{c}{\locattn{}} & \multicolumn{1}{c}{\contattn{}} \\
        SRN & \multicolumn{1}{c}{\cellcolor{\noncolor}} & 0.43 & 0.64 \\
        GRU & 0.01 & 0.78 & 0.17 \\
        LSTM & 0.02 & 0.05 & 0.01 \\
    \end{tabular}
    \subcaption{First-word accuracy on the generalization set}
    \end{subfigure}
    
    \caption{Results for each combination of recurrent unit and attention type. All numbers are medians over 100 initializations. 
    \noattn{} = no attention; \locattn{} = location-based attention; \contattn{} = content-based attention.
    A grayed-out cell indicates that the architecture scored below 50\% on the test set. In (b), the SRN \locattn{} produced the first auxiliary 45\% of the time; for all other models, the proportion of first-auxiliary outputs is almost exactly one minus the first-word accuracy (i.e., the proportion of main-auxiliary outputs).
    }
    \label{tab:architectures_final}
\end{figure*}

\subsection{Overview of experiments}

Holding the task constant, we first varied two aspects of the architecture that have no clear connection to question formation, namely the recurrent unit and the type of attention; both of these aspects have been central to major advances in natural language processing \citep{sundermeyer2012lstm,bahdanau2015}, so we investigate them here to see whether their contributions might be partially explained by linguistically-relevant inductive biases that they impart. 
We also tested a more clearly task-relevant modification of the architecture, namely the use of tree-based models rather than the sequential structure 
in Figure \ref{fig:seq2seq}.\footnote{Our code is at \url{github.com/tommccoy1/rnn-hierarchical-biases}. Results for the over 3500 models trained for this paper, with example outputs, are at \url{rtmccoy.com/rnn_hierarchical_biases.html}; only aggregate (median) results are reported in this paper.\label{fn:website}}

\section{Recurrent unit and attention}

\subsection{Recurrent unit}

The recurrent unit is the component that updates the hidden state after each word for the encoder and decoder. 
We used three types of recurrent units: simple recurrent networks (SRNs; \citealp{elman1990}), gated recurrent units (GRUs; \citealp{cho2014}), and long short-term memory (LSTM) units \citep{hochreiter1997}.
In SRNs and GRUs, the hidden state is represented by a single vector, while LSTMs use two vectors (the hidden state and the cell state). In addition, GRUs and LSTMs both use \textit{gates}, which control what information is retained across time steps, while SRNs do not; GRUs and LSTMs differ from each other in the number and types of gates they use.

\subsection{Attention}

In the basic model in Figure \ref{fig:seq2seq}, the final hidden state of the encoder is the decoder's only source of information about the input. 
To avoid having such a bottleneck,
many contemporary sequence-to-sequence models use \textbf{attention} \cite{bahdanau2015},
a feature that enables the decoder to consider all encoder hidden states ($E_0$ through $E_6$ in Figure \ref{fig:seq2seq}) when generating hidden state $D_i$. 
A model without attention has the only inputs to $D_i$ being $D_{i-1}$ and $y_{i-1}$ (the previous output); attention adds a third input, $c_i = \sum_j \alpha_i[j] E_j$, which is a weighted sum of the encoder's hidden states ($E_0$ through $E_n$) using a weight vector $\alpha_i$ whose $j^{th}$ element is denoted by $\alpha_i[j]$.

\begin{figure*}[]
    \begin{subfigure}{0.5\textwidth}
    \centering
    \begin{tabular}{cFF}
         & \multicolumn{1}{p{1.7cm}}{Unsquashed} & \multicolumn{1}{c}{Squashed} \\
        GRU \locattn{} & 0.99 & 0.77 \\
        LSTM \locattn{} & 0.98 & 0.98 \\
    \end{tabular}
    \subcaption{Full-sentence accuracy on the test set}
    \end{subfigure}
    \begin{subfigure}{0.5\textwidth}
    \centering
    \begin{tabular}{cHH}
         & \multicolumn{1}{p{1.7cm}}{Unsquashed} & \multicolumn{1}{c}{Squashed} \\
        GRU \locattn{} & 0.54 & 0.78 \\
        LSTM \locattn{} & 0.05 & 0.43 \\
    \end{tabular}
    \subcaption{First-word accuracy on the generalization set}
    \end{subfigure}
    \caption{Effects of squashing.
    All numbers are medians across 100 initializations.
    The standard versions of the architectures are the squashed GRU and the unsquashed LSTM. 
    }
    \label{tab:squeeze}
\end{figure*}

Implementations of attention vary in how the weights $\alpha_i[j]$ are derived \cite{graves2014neural,luong2015attention,chorowski2015attention}. 
Attention can be solely \textit{location-based}, where each $\alpha_i$ is determined solely from $D_{i-1}$ (and potentially also $y_{i-1}$), so that the model chooses \textit{where} to attend without first checking \textit{what} it is attending to.
Alternately, attention could be \textit{content-based},
in which case each $\alpha_i[j]$ is determined from both $D_{i-1}$ and $E_j$, such that the model does consider what it might attend to before attending to it. We test both location-based and content-based attention, and we also test models without attention.

\subsection{Results} \label{sec:archresults}

We trained models with all nine possible combinations of recurrent unit and attention type, 
using the hyperparameters and training procedure described in Appendix \ref{app:details}. 
The results are in Figure \ref{tab:architectures_final}.

The SRN without attention failed on the test set, mainly because it often confused words that had the same part of speech,
a known weakness of SRNs \cite{frank2007}. Therefore, its generalization set behavior is uninformative. 
The other architectures performed strongly on the test set 
($>$~50\% full-sentence accuracy), 
so we now consider their generalization set performance. The GRU with location-based attention and the SRN with content-based attention both preferred \textsc{move-main}, while the remaining architectures preferred \textsc{move-first}.\footnote{We say that a model \textit{preferred} generalization A over generalization B if it behaved more consistently with A than B.} 
These results suggest that both the recurrent unit and the type of attention can qualitatively affect a model's inductive biases. 
Moreover, the \textit{interactions} of these factors can have drastic effects: with SRNs, content-based attention led to behavior consistent with \textsc{move-main} while location-based attention led to behavior consistent with \textsc{move-first}; these types of attention had opposite effects with GRUs.

\subsection{Differences between LSTMs and GRUs}

One striking result in Figure \ref{tab:architectures_final} is that LSTMs and GRUs display qualitative differences, even though the two architectures are often viewed as interchangeable and  achieve similar performance in applied tasks \cite{chung2014empirical}.
One difference between LSTMs and GRUs is that a squashing function is applied to the hidden state of a GRU to keep its values within the range $(-1,1)$, while the cell state of an LSTM is not bounded. \newcite{weiss2018} demonstrate that such squashing leads to a qualitative difference in how well these models generalize counting behavior. 
Such squashing may also explain the qualitative differences that we observe:
counting the input elements is equivalent to keeping track of their linear positions, so we might expect that a tendency to count would make the linear generalization more accessible.

To test whether squashing increases a model's preference for \textsc{move-main}, we created a modified LSTM that included squashing in the calculation of its cell state,
and a modified GRU that did not have the squashing usually present in GRUs. See Appendix \ref{app:squashing} for more details.
Using the same training setup as before, we trained models with these modified recurrent units and with location-based attention. LSTMs and GRUs with squashing chose \textsc{move-main} more often than the corresponding models without squashing (Figure \ref{tab:squeeze}), suggesting that such squashing is one factor that causes GRUs to behave differently than LSTMs.

\subsection{Hyperparameters and random seed} \label{sec:init}

In addition to variation across architectures, we also observed considerable variation across multiple instances of the same architecture that differed only in random seed; the random seeds determined both the initial weights of each model and the order in which training examples were sampled.
For example, the generalization set first-word accuracy for SRNs with content-based attention ranged from 0.17 to 0.90.
Based on our exploration of hyperparameters, it also appears that the learning rate and hidden size can qualitatively affect generalization. The effects of these details are difficult to interpret systematically, and we leave the characterization of their effects for future work. Results for all individual re-runs 
are at the project website.\cref{fn:website}

\section{Tree models} \label{sec:tree}

So far we have tested whether properties that are not interpretably related to hierarchical structure nevertheless affect how a model generalizes on a syntactic task. 
We now turn to a related but opposite question: when a model's design is meant to give it a hierarchical inductive bias, does this design succeed at giving the model this bias?

\subsection{Tree model that learns implicit structure}

The first hierarchical model that we test is the Ordered Neurons LSTM (ON-LSTM; \citealp{shen2018ordered}).
This model is not given the tree structure of each sentence as part of its input. Instead, its processing is structured in a way that leads to the implicit construction of a soft parse tree.
This implicit tree structure is created by imposing a stack-like constraint on the updates to the values in the cell state of an LSTM: the degree to which the $i^{\textrm{th}}$ value is updated must always be less than or equal to the degree to which the $j^{\textrm{th}}$ value is updated for all $j \leq i$. This hierarchy of cell-state values adds an implicit tree structure to the model, where each level in the tree is defined by a soft depth in the cell state to which that level extends. 

We re-implemented the ON-LSTM and trained 100 instances of it using the hyperparameters specified in Appendix \ref{app:details}. 
This model achieved a test set full-sentence accuracy of 0.93 but a generalization set first-word accuracy of 0.05, showing a strong preference for \textsc{move-first} over \textsc{move-main}, contrary to what one would expect from a model with a hierarchical inductive bias. This lack of hierarchical behavior might be explained by \citeauthor{dyer2019onlstm}'s (\citeyear{dyer2019onlstm}) finding that ON-LSTMs do not perform much better than standard LSTMs at implicitly recovering hierarchical structure, even though ON-LSTMs (but not standard LSTMs) were designed in a way intended to impart a hierarchical bias. According to \citeauthor{dyer2019onlstm}, the \mbox{ON-LSTM's} apparent success reported in \newcite{shen2018ordered} was largely due to the method used to analyze the model rather than the model itself.

\begin{figure}
    \centering
    \includegraphics[width=0.7\columnwidth]{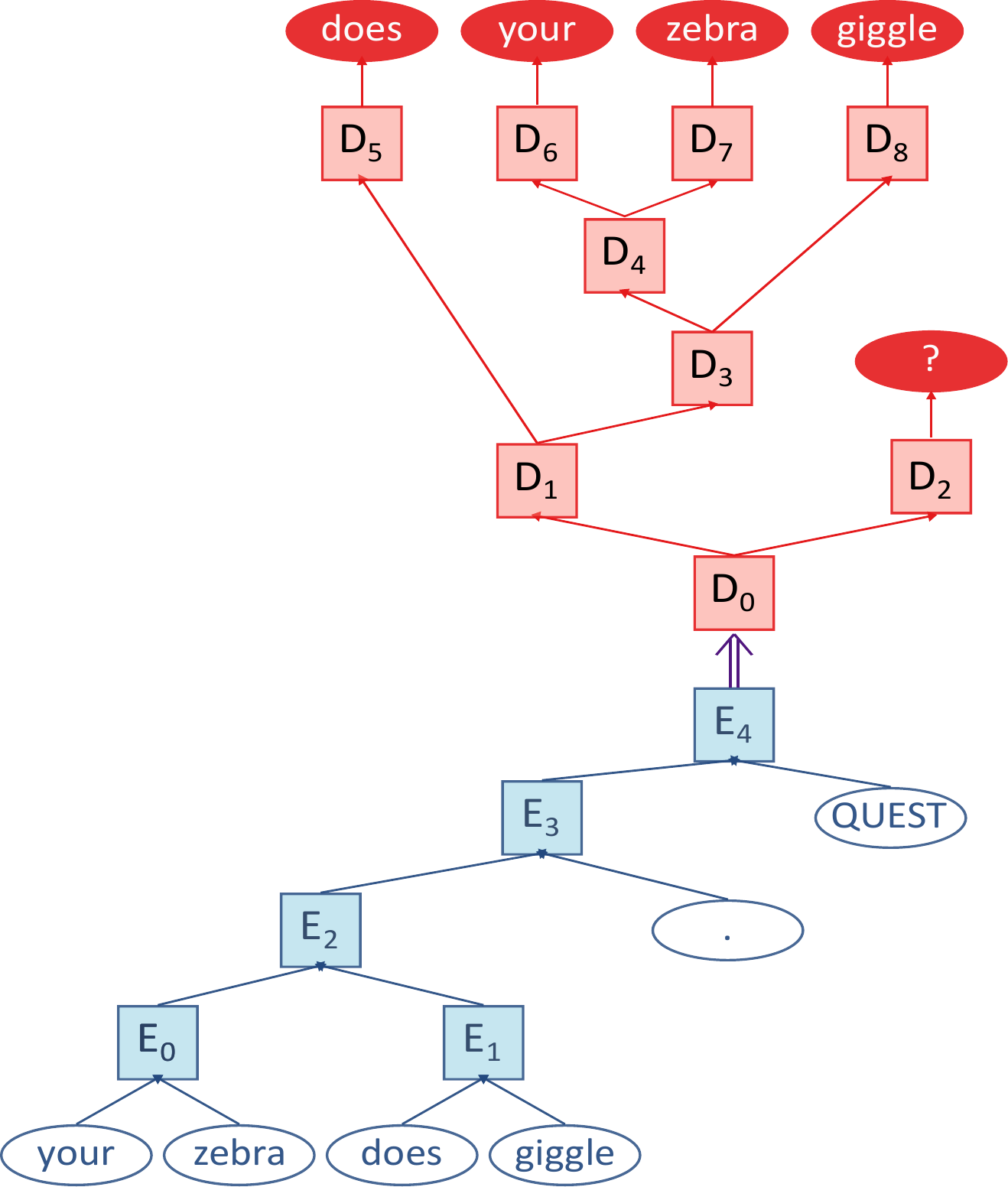}
    \caption{Sequence-to-sequence network with a tree-based encoder and tree-based decoder.}
    \label{fig:treernn}
\end{figure}

\subsection{Tree models given explicit structure}

The ON-LSTM results show that hierarchically structured processing alone is not sufficient to induce a bias for \textsc{move-main}, suggesting that constraints on \textit{which} trees are used may also be necessary.
We therefore tested a second type of hierarchical model, namely Tree-RNNs, that were explicitly fed the correct parse tree.
Parse trees can be used to guide the encoder, the decoder, or both; Figure~\ref{fig:treernn} shows a model where both the encoder and decoder are tree-based. For the tree-based encoder, we use the Tree-GRU from \newcite{chen2017improved}. This model composes the vector representations for a pair of sister nodes to generate a vector representing their parent.
It performs this composition bottom-up, starting with the word embeddings at the leaves and ending with a single vector representing the root ($E_4$ in Figure \ref{fig:treernn}); this vector acts as the encoding of the input.
For the tree-based decoder, we use a model based on the Tree-LSTM decoder from \citet{chen2018tree}, but using a GRU instead of an LSTM, for consistency with the tree encoder. 
This tree decoder is the mirror image of the tree encoder: starting with the vector representation of the root node ($D_0$ in Figure~\ref{fig:treernn}), it takes the vector representation of a parent node and outputs two vectors, one for the left child and one for the right child, until it reaches a leaf node, 
where it outputs a word.
We test models with a tree-based encoder and sequential decoder, a sequential encoder and tree-based decoder, or a tree-based encoder and tree-based decoder, all without attention; we investigate these variations to determine whether hierarchical generalization is determined by the encoder, the decoder, or both.

The results for these models are in Figure~\ref{tab:trees}, along with the previous results of the fully sequential GRU (sequential encoder + sequential decoder) without attention for comparison. 
The model with a tree-based encoder and sequential decoder preferred \textsc{move-first}, like the fully sequential model.
Only the models with a tree-based decoder preferred \textsc{move-main}, consistent with the finding of \newcite{mccoy2018rnns} that it is the decoder that determines an encoder-decoder model's representations. However, the model with a sequential encoder and a tree decoder failed on the test set, so the only model that both succeeded on the test set and showed a bias toward a  \textsc{move-main} generalization was the fully tree-based model (Tree/Tree).\footnote{We do not have an explanation for the failure of the sequential/tree model on the test set; most of its errors involved confusion among words that had the same part of speech (e.g. generating \textit{my} instead of \textit{your}).}
The behavior of this Tree/Tree model was striking in another way as well: Its generalization set \textit{full-sentence} accuracy was 69\%, while all other models---even those that achieved high \textit{first-word} accuracy on the generalization set---had close to 0\% generalization set full-sentence accuracy.
The ON-LSTM and Tree-GRU results show that an architecture designed to have a certain inductive bias might, but will not necessarily, display the intended bias.

\newcolumntype{x}[1]{>{\centering\arraybackslash\hspace{0pt}}p{#1}}

\begin{figure}[]
    \centering
    \resizebox{\columnwidth}{!}{
    \begin{tabular}{lYZ} \toprule
        \multicolumn{1}{c}{Model} & \multicolumn{1}{x{2.1cm}}{ Full-sentence test acc.} & \multicolumn{1}{x{2.1cm}}{First-word gen. acc.} \\ \midrule
        Sequential/Sequential & 0.88 & 0.03 \\
        Sequential/Tree & 0.00 & 0.90 \\
        Tree/Sequential & 0.96 & 0.13 \\
        Tree/Tree & 0.96 & 0.99 \\ \bottomrule
    \end{tabular}
    }
    \caption{Results with tree-based models (medians over 100 initializations). Model names indicate encoder/decoder; e.g., Sequential/Tree has a sequential GRU encoder and a tree-GRU decoder. 
    }
    \label{tab:trees}
\end{figure}

\section{Tense reinflection}

We have shown that several models reliably preferred \textsc{move-main} over \textsc{move-first}. However, this behavior alone does not necessarily mean that these models have a hierarchical bias, because a preference for \textsc{move-main} might arise not from a hierarchical bias but rather from some task-specific factors such as the prevalence of certain n-grams \cite{kam2008,berwick2011poverty}. A true hierarchical bias would lead a model to adopt hierarchical generalizations across training tasks; by contrast, we hypothesize that other factors (such as a bias for focusing on n-gram statistics) will be more sensitive to details of the task and will thus be unlikely to consistently produce hierarchical preferences. 
To test the robustness of the hierarchical preferences of our models, then, we introduce a second task, \textbf{tense reinflection}.

\subsection{Reinflection task} \label{sec:reinflect}

The reinflection task uses English subject-verb agreement to illuminate a model's syntactic generalizations \cite{linzen2016assessing}. 
The model is fed a past-tense English sentence as input. It must then output that sentence either unchanged or transformed to the present tense, with the final word of the input indicating the task to be performed:

\ex. \label{ex:past} my yak  swam . \textsc{past} $\rightarrow$ my yak swam . 

\ex. \label{ex:present}my yak  swam . \textsc{present} $\rightarrow$ my yak swims . 

\noindent
Because the past tense in English does not inflect for number (e.g., the past tense of \textit{swim} is \textit{swam} whether the subject is singular or plural), the model must determine from context whether each verb
being turned to present tense 
should be singular or plural. Example \ref{ex:present} is consistent with two salient rules for determining which aspects of the context are relevant:

\ex. \textsc{agree-subject}: Each verb should agree with its hierarchically-determined subject.

\ex. \textsc{agree-recent}:  Each verb should agree with the linearly most recent noun.

\noindent
Though these rules make the same prediction for \ref{ex:present}, they make different predictions for other examples, such as \ref{ex:tensedisamba}, for which \textsc{agree-subject} predicts \ref{ex:tensedisambb} while \textsc{agree-recent} predicts \ref{ex:tensedisambc}:

\ex. \a. my zebra by the yaks swam . \textsc{present} \label{ex:tensedisamba}
\b. my \textbf{zebra} by the yaks \textbf{swims} . \label{ex:tensedisambb}
\c. my zebra by the \textbf{yaks} \textbf{swim} . \label{ex:tensedisambc}

\noindent 
Similar to the setup for the question formation experiments, we trained models on examples for which \textsc{agree-subject} and \textsc{agree-recent} made the same predictions and evaluated the trained models on examples where the rules make different predictions. 
 We ran this experiment with all 9 sequential models ([SRN, GRU, LSTM] x [no attention, location-based attention, content-based attention]), the ON-LSTM, and the model with a tree-based encoder and tree-based decoder that were provided the correct parse trees, using the hyperparameters in Appendix \ref{app:details}. The example sentences were generated using the same context-free grammar used for the question formation task, except with inflected verbs instead of auxiliary/verb bigrams (e.g., \textit{reads} instead of \textit{does read}). We evaluated these models on the full-sentence accuracy on the test set and also main-verb accuracy for the generalization set---that is, the proportion of generalization set examples for which the main verb was correctly predicted, such as when \textit{swims} rather than \textit{swim} was chosen in the output for \ref{ex:tensedisamba}. Models usually chose the correct lemma for the main verb (at least 87\% of the time for all tense reinflection models), with most main verb errors involving the correct verb but with incorrect inflection (i.e., being singular instead of plural, or vice versa).
Thus, a low main-verb accuracy can be interpreted as consistency with \textsc{agree-recent}.

\begin{figure}[]
    \centering
    \resizebox{\columnwidth}{!}{
    \begin{tabular}{lYZ} \toprule
        \multicolumn{1}{c}{Model} & \multicolumn{1}{p{2.1cm}}{ Full-sentence test acc.}  & \multicolumn{1}{p{2.1cm}}{Main-verb gen. acc.} \\ \midrule
        SRN \noattn{} & 0.00 & \multicolumn{1}{c}{\cellcolor{\noncolor}}  \\
        SRN \locattn{} & 1.00 & 0.00 \\
        SRN \contattn{} & 1.00 & 0.00 \\
        GRU \noattn{} & 0.90 & 0.04 \\
        GRU \locattn{} & 0.81 & 0.00 \\
        GRU \contattn{} & 1.00 & 0.00 \\
        LSTM \noattn{} & 0.96 & 0.04 \\
        LSTM \locattn{} & 0.98 & 0.00 \\
        LSTM \contattn{} & 1.00 & 0.00\\
        ON-LSTM \noattn{} & 0.95 & 0.05 \\
        Tree/Tree \noattn{} & 0.96 & 0.94  \\ \bottomrule
    \end{tabular}
    }
    \caption{Reinflection results (medians over 100 initializations). \noattn{} = no attention; \locattn{} = location-based attention; \contattn{} = content-based attention.}
    \label{tab:reinflection}
\end{figure}

All sequential models, even the ones that generalized hierarchically with question formation, overwhelmingly chose \textsc{agree-recent} for this reinflection task (Figure~\ref{tab:reinflection}), consistent with the results of a similar experiment done by \newcite{ravfogel2019studying}. The ON-LSTM also preferred  \textsc{agree-recent}.
By contrast, the fully tree-based model preferred the hierarchical generalization \textsc{agree-subject}. Thus, although the question formation experiments showed qualitative differences in sequential models' inductive biases, this experiment shows that those differences cannot be explained by positing that there is a general hierarchical bias in some of our sequential models.
What the relevant bias for these models \textit{is} remains unclear; we only claim 
to show that it is not a hierarchical bias. 
Overall, the model with both a tree-based encoder and a tree-based decoder is the only model we tested that plausibly has a generic hierarchical bias, as it is the only one that behaved consistently with such a bias across both tasks.

\section{Are tree models constrained to generalize hierarchically?}

It may seem that the tree-based models are constrained by their structure to make only hierarchical generalizations, rendering their hierarchical generalization trivial.
In this section, we test whether they are in fact constrained in this way, and similarly  whether sequential models are constrained to make only linear generalizations.
Earlier, the training sets for our two tasks were ambiguous between two generalizations, but we now used training sets that unambiguously supported either a linear transformation or a hierarchical transformation.\footnote{The lack of ambiguity in each training set means that the generalization set becomes essentially another test set. 
} 
For example, we used a \textsc{move-main} training set that included some examples like \ref{ex:movemain}, while the \textsc{move-first} training set included some examples like \ref{ex:movefirst}:

\ex. \a. my yaks that do read don't giggle . \textsc{quest} $\rightarrow$ don't my yaks that do read giggle ? \label{ex:movemain}
\b. my yaks that do read don't giggle . \textsc{quest} $\rightarrow$ do my yaks that read don't giggle ? \label{ex:movefirst}

Similarly, for the tense reinflection task, we created an \textsc{agree-subject} training set and an \textsc{agree-recent} training set. For each of these four training sets, we trained 100 sequential GRUs and 100 Tree/Tree GRUs, all without attention.

Each model learned to perform linear and hierarchical transformations with similar accuracy:
On the \textsc{move-main} and \textsc{move-first} datasets, both the sequential and tree-based models achieved 100\% first-word accuracy. On both the \textsc{agree-subject} and \textsc{agree-recent} datasets, the sequential model achieved 91\% main-verb accuracy and the tree-based model achieved 99\% main-verb accuracy.
Thus, the fact that the tree-based model preferred hierarchical generalizations when the training set was ambiguous arose not from any constraint imposed by the tree structure but rather from the model's inductive biases---biases that can be overridden given appropriate training data.

\section{Tree structure vs.\ tree information}

Our sequential and tree-based models differ not only in structure but also in the information they have been provided: the tree-based models have been given correct parse trees for their input and output sentences, while the sequential models have not been given parse information. Therefore, it is unclear whether the hierarchical generalization displayed by the tree-based models arose from the tree-based model structure, from the parse information provided to the models, or both.

To disentangle these factors, we ran two further experiments. First, we retrained the Tree/Tree GRU but using uniformly right-branching trees (as in (11b)) instead of correct parses (as in (11a)).
Thus, these models make use of tree structure but not the kind of parse structure that captures linguistic information. Second, we retrained the sequential GRU without attention\footnote{We chose this sequential model because the Tree/Tree model is also based on GRUs without attention.} but modified the input and output by adding brackets that indicate each sentence's parse; for example, \ref{ex:plain} would be changed to \ref{ex:bracket}. Thus, these models are provided with parse information in the input but such structure does not guide the neural network computation as it does with tree RNNs.

\ex. a.  \resizebox{0.19\textwidth}{!}{\Tree [ [ [ my yak ] [ does giggle ] ]  $.$ ]} \hfill b.  \resizebox{0.19\textwidth}{!}{\Tree [ my [ yak  [ does  [ giggle  $.$ ] ] ] ]}

\ex. \a. my yak does giggle . \textsc{quest} \\$\rightarrow$  does my yak giggle ? \label{ex:plain}
\b. [ [ [ my yak ] [ does giggle ]  . ] \textsc{quest} ] $\rightarrow$ [ [ does [ [ my yak ] giggle ] ] ? ] \label{ex:bracket}

We ran 100 instances of each experiment using different random seeds. For the experiment with bracketed input, the brackets significantly increased the lengths of the sentences, making the learning task harder; we therefore found it necessary to use a patience of 6 instead of the patience of 3 we used elsewhere, but all other hyperparameters remained as described in Appendix \ref{app:details}.

\begin{figure}[h]
\begin{subfigure}{\columnwidth}
    \centering
    \begin{tabular}{cZZ} \toprule
         & \multicolumn{1}{c}{Not provided}  & \multicolumn{1}{c}{Provided}  \\
         & \multicolumn{1}{c}{correct parse} & \multicolumn{1}{c}{correct parse} \\ \midrule
        GRU \noattn{} & 0.01 & 0.35 \\
        Tree/Tree \noattn{} & 0.05  & 0.99 \\ \bottomrule
    \end{tabular}
    \caption{Question formation generalization set results.}
    \label{tab:ablationsubjaux}
\end{subfigure}
\vspace{\baselineskip}
\\
\begin{subfigure}{\columnwidth}
    \centering
    \begin{tabular}{cZZ} \toprule
         & \multicolumn{1}{c}{Not provided}  & \multicolumn{1}{c}{Provided}  \\
         & \multicolumn{1}{c}{correct parse} & \multicolumn{1}{c}{correct parse} \\ \midrule
        GRU \noattn{} & 0.04 & 0.00 \\
        Tree/Tree \noattn{} & 0.07  & 0.94 \\ \bottomrule
    \end{tabular}
    \caption{Tense reinflection generalization set results.}
    \label{tab:ablationtense}
\end{subfigure}
\caption{Disentangling tree structure and parse information. The GRU \noattn{} that is not provided the correct parse is the same as GRU \noattn{} in Figures \ref{tab:architectures_final} and \ref{tab:reinflection}. The Tree/Tree model that is provided the correct parse is the same as the Tree/Tree model in Figures \ref{tab:trees} and \ref{tab:reinflection}. The other two conditions are new: The GRU \noattn{} that was provided the correct parses was given these parses via bracketing, while the Tree/Tree model that was not provided the correct parses was instead given right-branching trees.}
\label{fig:ablation}
\end{figure}

For both tasks, neither the sequential GRU that was given brackets in its input nor the Tree/Tree model that was given right-branching trees displayed a hierarchical bias (Figure \ref{fig:ablation}).\footnote{Providing the parse with brackets did significantly improve the first-word accuracy of the sequential GRU, but this accuracy remained below 50\%.}
The lack of hierarchical bias in the sequential GRU with bracketed input indicates that simply providing parse information in the input and target output is insufficient to induce a model to favor hierarchical generalization; it appears that such parse information must be integrated into the model's structure to be effective. On the other hand, the lack of a hierarchical bias in the Tree/Tree model using right-branching trees shows that simply having tree structure is also insufficient; it is necessary to have the \textit{correct} tree structure.

\section{Will models generalize across transformations?}

Each experiment discussed so far involved a single linguistic transformation. By contrast, humans acquiring language are not exposed to phenomena in isolation but rather to a complete language encompassing many phenomena. This fact has been pointed to as a possible way to explain hierarchical generalization in humans without needing to postulate any innate preference for hierarchical structure. 
While one phenomenon, such as question formation, might be ambiguous in the input, there might be enough direct evidence among other phenomena to conclude that the language as a whole is hierarchical, a fact which learners can then extend to the ambiguous phenomenon \cite{pullum2002empirical,perfors2011learnability}, under the non-trivial assumption that the learner will choose to treat the disparate phenomena in a unified fashion.

While our training sets are ambiguous with respect to whether the phenomenon underlying the mapping is structurally driven, they do contain other cues that the language is more generally governed by hierarchical regularities. First, certain structural units are reused across positions in a sentence; for example, prepositional phrases can appear next to subjects or objects. Such reuse of structure can be represented more efficiently with a hierarchical grammar than a linear one. 
Second, in the question formation task, subject-verb agreement can also act as a cue to hierarchical structure: e.g., in the sentence \textit{my \textbf{walrus} by the yaks \textbf{does} read}, the inflection of \textit{does} depends on the verb's hierarchically-determined subject (\textit{walrus}) rather than the linearly closest noun (\textit{yaks}).\footnote{Subject-verb agreement does not act as a cue to hierarchy in the tense reinflection task because 
all relevant sentences have been withheld
to maintain the training set's ambiguity.
}

For the sequential RNNs we have investigated, it appears that these indirect cues to hierarchical structure were not sufficient to guide the models towards hierarchical generalizations. However, perhaps the inclusion of some more direct evidence for hierarchy would be more successful.  

\begin{figure}[]
    \centering
    \begin{tabular}{p{2cm}ZZ} \toprule
        & \multicolumn{1}{c}{Ambiguous} & \multicolumn{1}{c}{Ambiguous}  \\
        & \multicolumn{1}{c}{question} & \multicolumn{1}{c}{tense}  \\
        & \multicolumn{1}{c}{ formation} & \multicolumn{1}{c}{reinflection}  \\ \midrule
        Single-task & 0.01 & 0.04 \\
        Multi-task  & 0.09 & 0.17 \\
        \parindent=1em   +auxiliaries  & 0.01 & 0.99 \\ \bottomrule
    \end{tabular}
    \caption{Multi-task learning results for a GRU without attention. \textit{Single-task} reports baselines from training on a single ambiguous task. \textit{Multi-task} reports results from adding an unambiguous second task. \textit{Multi-task + auxiliaries} reports results from adding an unambiguous second task and also adding overt auxiliaries to the tense reinflection sentences. 
    The numbers give the generalization set performance on the ambiguous task.
    }
    \label{tab:multitask}
\end{figure}

To take a first step toward investigating this possibility, we use a multi-task learning setup, where we train a single model to perform both question formation and tense reinflection. We set up the training set such that one task was unambiguously hierarchical while the other was ambiguous between the hierarchical generalization and the linear generalization. This gave two settings: One where question formation was ambiguous, and one where tense reinflection was ambiguous. We trained 100 instances of a GRU without attention on each setting and assessed how each model generalized for the task that was ambiguous.

For both cases, generalization behavior in the multi-task setting differed only minimally from the single-task setting (Figure \ref{tab:multitask}). 
One potential explanation for the lack of transfer across tasks is that the two tasks operated over different sentence structures: the question formation sentences always contained overt auxiliaries on their verbs (e.g., \textit{my walrus does giggle}), while the tense reinflection sentences did not (e.g., \textit{my walrus giggles}). To test this possibility, we reran the multi-task experiments but with overt auxiliaries added to the tense reinflection sentences (Figure~\ref{tab:multitask}, ``Multi-task + auxiliaries'' row). In this setting, the model still generalized linearly when it was question formation that was ambiguous.
However, when it was tense reinflection that was ambiguous, the model generalized hierarchically.

We hypothesize that the directionality of this transfer is due to the fact that the question formation training set includes unambiguous long-distance subject-verb agreement as in \ref{ex:questagr}, which might help the model on generalization-set examples for tense reinflection such as  \ref{ex:tenseagr}: 

\ex. my  \textbf{zebras} by the yak \textbf{do} read . \textsc{decl} \newline $\rightarrow$ my \textbf{zebras} by the yak \textbf{do} read . \label{ex:questagr}

\ex. my zebras by the yak did read .  \textsc{present} \newline $\rightarrow$ my \textbf{zebras} by the yak \textbf{do} read . \label{ex:tenseagr}

By contrast, the tense reinflection training set does not contain any outputs of the type withheld from the question formation training set. 
If this explanation is correct, it would mean that the improvement on the tense reinflection task 
derived not from the question formation \textit{transformation} but rather from the subject-verb agreement incidentally present in the question formation \textit{dataset}. Therefore, even the single potential case of generalization across transformations is likely spurious.

Recent NLP work has also found that neural networks do not readily transfer knowledge across tasks; e.g., pretrained models often perform worse than non-pretrained models 
\cite{wang2019tell}. This lack of generalization across tasks might be due to the tendency of multi-task neural networks to create largely independent representations for different tasks even when a shared representation could be used \cite{kirov2012processing}.
Therefore, to make cross-phenomenon generalizations, neural networks may need to be given an explicit bias for sharing processing across phenomena.

\section{Discussion}

We have found that all factors we tested can qualitatively affect a model's inductive biases but that a hierarchical bias---which has been argued to underlie children's acquisition of syntax---only arose in a model whose inputs and computations were governed by  syntactic structure. 

\subsection{Relation to \textit{Rethinking Innateness}}

Our experiments were motivated in part by the book \textit{Rethinking Innateness} \cite{elman1998rethinking} which argued that humans' inductive biases must arise from constraints on the wiring patterns of the brain.
Our results support two conclusions from this book. First, those authors argued that ``Dramatic effects can be produced by small changes'' (p. 359). This claim is supported by our observation that low-level factors, such as the size of the hidden state, qualitatively affect how models generalize (Section \ref{sec:init}). 
Second, they argued that ``[w]hat appear to be single events or behaviors may have a multiplicity of underlying causes'' (p. 359); in our case, we found that a model's generalization behavior results from some combination of factors that interact in hard-to-interpret ways; e.g., changing the type of attention had different effects in SRNs than in GRUs.

The dramatic effects of these low-level factors offer some support for the claim that humans' inductive biases can arise from fine-grained architectural constraints in the brain. However, this support is only partial.
Our only model that robustly displayed the kind of preference for hierarchical generalization that is necessary for language learning did not derive such a preference from low-level architectural properties but rather from the explicit encoding of linguistic structure.

\subsection{Relation to human language acquisition}

Our experiments showed that some tree-based models displayed a hierarchical bias, while non-tree-based models never displayed such a bias, even when provided with strong cues to hierarchical structure in their input (through bracketing or multi-task learning).
These findings suggest that the hierarchical preference displayed by humans when acquiring English requires making explicit reference to hierachical structure, and cannot be argued to emerge from more general biases applied to input containing cues to hierarchical structure. 
Moreover, since the only successful hierarchical model was one that took the correct parse trees as input, our results suggest that a child's set of biases includes biases governing which specific trees will be learned.
Such biases could involve innate knowledge of likely tree structures, but they do not need to; they might instead involve
innate tendencies to bootstrap parse trees from other sources, such as prosody \citep{morgan1996signal} or semantics \citep{pinker1996language}.
With such information, children might learn their language's basic syntax before  beginning to acquire question formation, and this knowledge might then guide their acquisition of question formation.

There are three important caveats for extending our conclusions to humans. First, humans may have a stronger bias to share processing across phenomena than neural networks do, in which case multi-task learning would be a viable explanation for the biases displayed by humans even though it had little effect on our models.
Indeed, this sort of cross-phenomenon consistency is similar in spirit to the principle of systematicity, and it has long been argued that humans have a strong bias for systematicity while neural networks do not \citep[e.g.,][]{fodor1988connectionism,lake2018generalization}.
Second, some have argued that children's input actually does contain utterances unambiguously supporting a hierarchical transformation \citep{pullum2002empirical}, whereas we have assumed a complete lack of such examples.
Finally, our training data omit many cues to hierarchical structure that are available to children, including prosody and real-world grounding. It is possible that, with data closer to a child's input, more general inductive biases might succeed. 

However, there is still significant value in studying what can be learned from strings alone, because we are unlikely to understand how the multiple components of a child's input interact without a better understanding of each component. Furthermore, during the acquisition of abstract aspects of language, real-world grounding is not always useful in the absence of linguistic biases \citep{gleitman1992picture}. 
More generally, it is easily possible for learning to be harder when there is more information available than when there is less information available \citep{dupoux2018cognitive}. 
Thus, our restricted experimental setup may actually make learning easier than in the more informationally-rich scenario faced by children.

\subsection{Practical takeaways}

Our results leave room for three possible approaches to imparting a model with a hierarchical bias. First, one could search the space of hyperparameters and random seeds to find a setting that leads to the desired generalization.
However, this may be ineffective: At least in our limited exploration of these factors, we did not find a hyperparameter setting that led to hierarchical generalization across tasks for any non-tree-based model.

A second option is to add a pre-training task or use multi-task learning \cite{caruana1997multitask, collobert2008unified, enguehard-etal-2017-exploring}, where the additional task is designed to highlight hierarchical structure. 
Most of our multi-task experiments only achieved modest improvements over the single-task setting, suggesting that this approach is also not very viable.
However, it is possible that further secondary tasks would bring further gains, making this approach more effective.

A final option is to use more interpretable architectures with explicit hierachical structure.
Our results suggest that this approach is the most viable, as it  yielded models that reliably generalized hierarchically. 
However, this approach only worked when the architectural bias was augmented with rich assumptions about the input to the learner, namely that it provided correct hierarchical parses for all sentences.
We leave for future work an investigation of how to effectively use tree-based models without providing correct parses.

\section*{Acknowledgments}

For helpful comments we thank Joe Pater, Paul Smolensky, the JHU Computation and Psycholinguistics lab, the JHU Neurosymbolic Computation lab, the Computational Linguistics at Yale (CLAY) lab, the anonymous reviewers, and audiences at the University of Pavia Center for Neurocognition, Epistemology, and Theoretical Syntax, the Penn State Dept. of Computer Science and Engineering, and the MIT Dept. of Brain and Cognitive Sciences. 
Any errors are our own.

This material is based upon work supported by the NSF Graduate Research
Fellowship Program under Grant No. 1746891, and by NSF Grant Nos. BCS-1920924 and BCS-1919321. Any opinions, findings, and conclusions or recommendations expressed in this material are
those of the authors and do not necessarily reflect the views of the National Science Foundation. Our experiments were conducted with resources from the Maryland Advanced Research Computing Center (MARCC). 

\appendix

\section{Architecture and training details} \label{app:details}

We used a word embedding size of 256 (with word embeddings learned from scratch), a hidden size of 256, a learning rate of 0.001, and a batch size of 5. Models were evaluated on a validation set after every 1000 training batches, and we halted training if the model had been trained for at least 30,000 batches and had shown no improvement over 3 consecutive evaluations on the validation set (the number 3 in this context is called the patience). The training set contained 100,000 examples, while the validation, test, and generalization sets contained 10,000 examples each. The datasets were held constant across experiments, but models sampled from the training set in different orders across experiments. During training, we used teacher forcing on 50\% of examples.

\section{Equations for squashing experiments} \label{app:squashing}

The equations governing a standard LSTM are:

\begin{align}
i_t &= \sigma(W_i * [h_{t-1}, w_{t}] + b_i) \\
f_t &= \sigma(W_f * [h_{t-1}, w_{t}] + b_f) \\
g_t &= \tanh(W_g * [h_{t-1}, w_{t}] + b_g) \\
o_t &= \sigma(W_o * [h_{t-1}, w_{t}] + b_o) \\
c_t &= f_t * c_{t-1} + i_t * g_t \label{eq:lstmcell}\\
h_t &= o_t * \tanh(c_t)
\end{align}
To create a new LSTM whose cell state exhibits squashing, like the hidden state of the GRU, we modified the LSTM cell state update in (\ref{eq:lstmcell}) to (\ref{eq:lstmsquashcell}), where the new coefficients now add to 1:\footnote{
We modified the structure of the gates rather than adding a squashing nonlinearity to avoid vanishing gradients.}

\begin{align}
c_t = \frac{f_t}{f_t + i_t} * c_{t-1} + \frac{i_t}{f_t + i_t} * g_t \label{eq:lstmsquashcell}
\end{align}

\noindent
The equations governing a standard GRU are:

\begin{align}
    r_t &= \sigma(W_r[h_{t-1}, w_{t}] + b_r) \\
    z_t &= \sigma(W_z[h_{t-1}, w_{t}] + b_z) \\
    \tilde{h} &= \tanh(W_x[r_t * h_{t-1}, w_t] + b_x) \\
    h_t &= z_t * h_{t-1} + (1 - z_t) * \tilde{h} \label{eq:gruhidden} 
\end{align}

\noindent
The GRU's hidden state is squashed because its update gate $z$ merges the functions of the input and forget gates ($i$ and $f$) of the LSTM 
(cf.\ equations \ref{eq:lstmcell} and \ref{eq:gruhidden}). 
As a result, the input and forget weights are  tied in the GRU but not the LSTM. To create a non-squashed GRU, we added an input gate $i$ and changed the hidden state update (Equation \ref{eq:gruhidden}) to Equation \ref{eq:hiddennew} to make $z$ act solely as a forget gate: 

\begin{align}
    i_t &= \sigma(W_i[h_{t-1}, w_{t}] + b_i) \\
    h_t &= z_t * h_{t-1} + i_t * \tilde{h} \label{eq:hiddennew}
\end{align}

\bibliography{tacl2018}
\bibliographystyle{acl_natbib}

\end{document}